\documentclass{article}

 \usepackage[preprint]{neurips_2025}

\usepackage[utf8]{inputenc} 
\usepackage[T1]{fontenc}    
\usepackage{hyperref}       
\usepackage{url}            
\usepackage{booktabs}       
\usepackage{amsfonts}       
\usepackage{nicefrac}       
\usepackage{microtype}      
\usepackage{xcolor}         

\usepackage{natbib}
\definecolor{darkblue}{rgb}{0, 0, 0.5}
\hypersetup{colorlinks=true, citecolor=darkblue, linkcolor=darkblue, urlcolor=darkblue}

\usepackage{graphicx}
\usepackage{fancyhdr}
\usepackage{xcolor}
\usepackage{etoolbox}

\usepackage{times}
\usepackage{latexsym}
\usepackage{inconsolata}
\usepackage{multirow}
\usepackage{graphicx}
\usepackage{amsmath}
\usepackage{url}
\usepackage{array}
\usepackage{makecell}
\usepackage{subcaption}
\usepackage{colortbl}
\usepackage{xcolor}
\usepackage{algorithm}
\usepackage{algpseudocode}

\usepackage{tikz}
\usepackage{fontawesome5}
\usepackage{hyperref}
\usepackage{xcolor}

\hypersetup{colorlinks=true, urlcolor=blue!70!black}

\algrenewcommand{\algorithmiccomment}[1]{\hfill\textcolor{gray}{\textit{// #1}}}
 
\definecolor{ours}{HTML}{D6EAF8}

\fancypagestyle{firstpage}{
  \fancyhead[L]{\includegraphics[height=2cm]{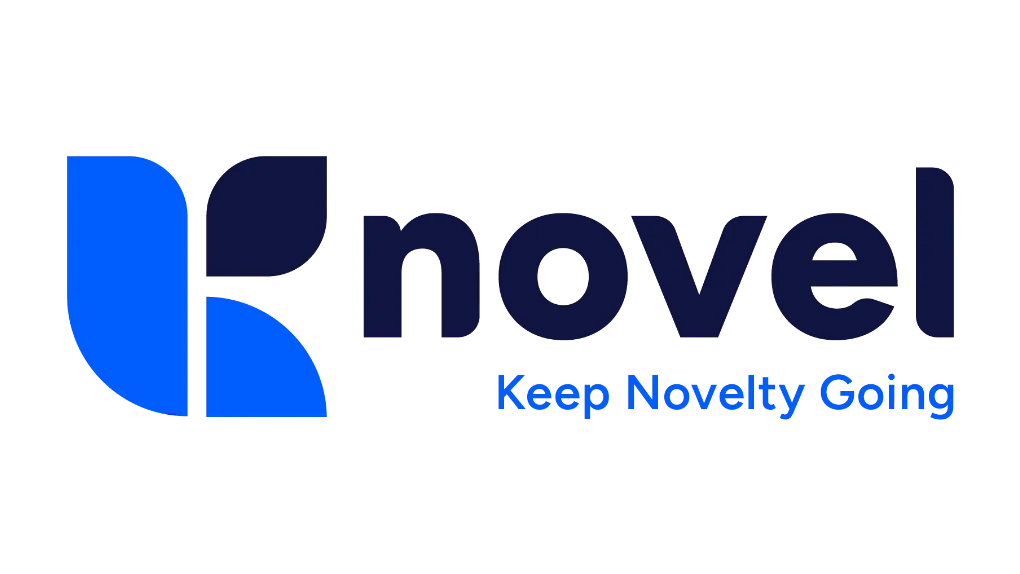}}
  \fancyhead[R]{}
  
}

\title{Polyglot-Lion: Efficient Multilingual ASR for Singapore via Balanced Fine-Tuning of Qwen3-ASR}

\author{%
  Quy-Anh Dang\thanks{Correspondence: quyanh.dang@knoveleng.com}, Chris Ngo \\
  Knovel Engineeing Lab, Singapore \\
  \texttt{\{quyanh.dang, chris.ngo\}@knoveleng.com} \\
}

\begin{document}

\maketitle
\thispagestyle{firstpage}

\begin{center}
\vspace{-0.5em}
\begin{tabular}{ccc}

  \href{https://knoveleng.github.io/polyglot-lion/}{%
    \fcolorbox{white}{white}{%
      \parbox{2.8cm}{\centering\vspace{4pt}
        {\large\color{black}\faGlobe}\\[3pt]
        \textbf{\small Project Page}%
      \vspace{4pt}}%
    }%
  }
  &\hspace{0.5cm}

  \href{https://github.com/knoveleng/polyglot-lion}{%
    \fcolorbox{white}{white}{%
      \parbox{2.8cm}{\centering\vspace{4pt}
        {\large\color{black}\faGithub}\\[3pt]
        \textbf{\small Code}%
      \vspace{4pt}}%
    }%
  }
  &\hspace{0.5cm}

  \href{https://huggingface.co/collections/knoveleng/polyglot-lion}{%
    \fcolorbox{white}{white}{%
      \parbox{2.8cm}{\centering\vspace{4pt}
        \includegraphics[height=1.1em]{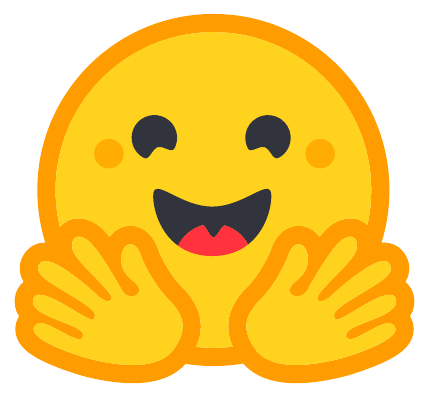}\\[3pt]
        \textbf{\small Models}%
      \vspace{4pt}}%
    }%
  }

\end{tabular}
\vspace{0.5em}
\end{center}

\begin{abstract}
We present \textbf{Polyglot-Lion}, a family of compact multilingual automatic speech recognition (ASR) models tailored for the linguistic landscape of Singapore, covering English, Mandarin, Tamil, and Malay. Our models are obtained by fine-tuning \texttt{Qwen3-ASR-0.6B} and \texttt{Qwen3-ASR-1.7B} exclusively on publicly available speech corpora, using a balanced sampling strategy that equalizes the number of training utterances per language and deliberately omits language-tag conditioning so that the model learns to identify languages implicitly from audio. On 12 benchmarks spanning the four target languages, \textbf{Polyglot-Lion-1.7B} achieves an average error rate of \textbf{14.85}, competitive with MERaLiON-2-10B-ASR (14.32) - a model 6$\times$ larger - while incurring a training cost of \textbf{\$81} on a single RTX PRO 6000 GPU compared to \textbf{\$18,862} for the 128-GPU baseline. Inference throughput is approximately \textbf{20$\times$ faster} than MERaLiON at 0.10 s/sample versus 2.02 s/sample. These results demonstrate that linguistically balanced fine-tuning of moderate-scale pretrained models can yield deployment-ready multilingual ASR at a fraction of the cost of larger specialist systems.
\end{abstract}

\section{Introduction}
\label{sec:intro}

Singapore presents a uniquely demanding setting for automatic speech recognition (ASR): four official languages - English, Mandarin Chinese, Tamil, and Malay - coexist in everyday communication, often within a single conversation or utterance. This linguistic landscape is further complicated by the prevalence of \textit{Singlish}, a creole variety that draws lexical and phonological material from all four languages, and by wide variation in speaker age, accent, and code-switching behaviour \citep{lim2004singapore}. Together, these factors make Singapore one of the most challenging real-world environments for multilingual ASR.

Despite this linguistic richness, high-quality open-source ASR systems that cover all four official languages simultaneously remain scarce. General-purpose multilingual models such as Whisper \citep{radford2023robust} and MMS \citep{pratap2023mms} provide broad language coverage through large-scale pretraining, but their accuracy degrades on lower-resource varieties such as Tamil and Malay and on Singapore-accented English \citep{koh19_interspeech_nsc}. Audio-language models (ALMs) such as Qwen2.5-Omni \citep{qwen25omni} and SeaLLMs-Audio \citep{seallmsaudio} extend speech recognition with general language understanding, yet their large parameter counts (7B+) render fine-tuning and deployment expensive. Specialist systems such as MERaLiON-2-10B-ASR \citep{meralion2} have been purpose-built for the Singapore multilingual setting and achieve strong performance across all four languages, but require 128 GPUs and an estimated \$18,862 to train - a barrier that places them beyond the reach of most academic groups and small enterprises.

\begin{figure}[!htbp]
    \centering
    \includegraphics[width=\linewidth]{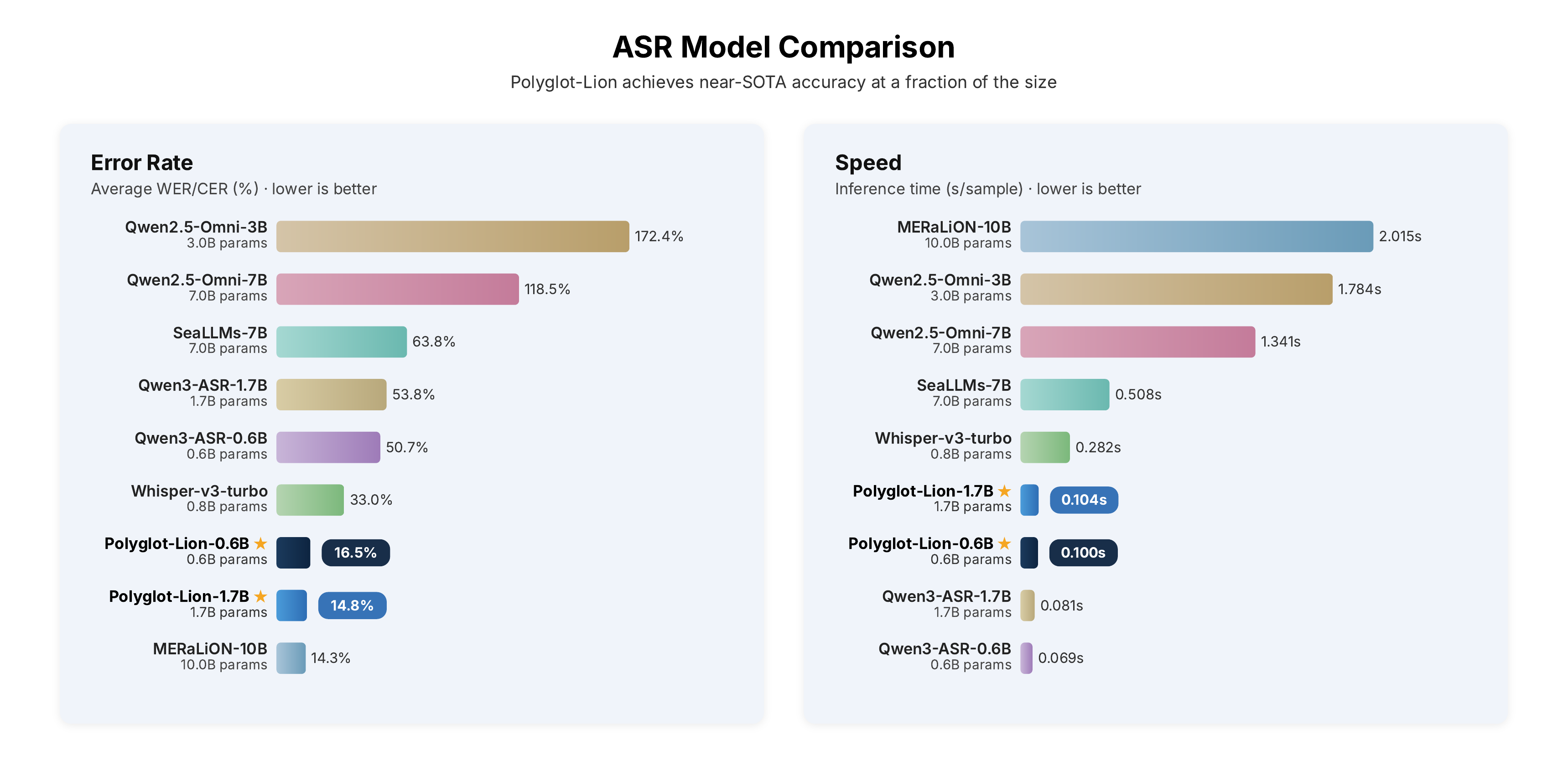}
    \caption{Polyglot-Lion achieves near-SOTA accuracy at a fraction of the model size and inference cost. \textbf{Left:} Average error rate (WER/CER) across 12 benchmarks; lower is better. \textbf{Right:} Inference speed in seconds per sample; lower is better. Despite having 6$\times$ fewer parameters than MERaLiON-2-10B-ASR, Polyglot-Lion-1.7B matches its accuracy while being approximately 20$\times$ faster at inference.}
    \label{fig:panel}
\end{figure}

In this paper, we introduce \textbf{Polyglot-Lion}\footnote{\url{https://github.com/knoveleng/polyglot-lion}} (\textit{Poly}: many; \textit{Glot}: tongue; \textit{Lion}: the lion-city, Singapore), a family of compact multilingual ASR models built by fine-tuning \texttt{Qwen3-ASR-0.6B} and \texttt{Qwen3-ASR-1.7B} \citep{qwen3asr} exclusively on publicly available speech corpora. As illustrated in Figure~\ref{fig:panel}, Polyglot-Lion-1.7B achieves an average error rate of 14.8 across 12 benchmarks - closely matching MERaLiON-2-10B-ASR (14.3) while running nearly 20$\times$ faster at inference time. This is accomplished through two simple but effective design choices: (1) a balanced sampling strategy that equalises per-language training coverage, and (2) the deliberate removal of language-tag conditioning, forcing the model to detect the spoken language directly from the acoustic signal.

Our contributions are as follows:

\begin{enumerate}
  \item \textbf{A balanced multilingual fine-tuning recipe} that upsamples under-represented languages to achieve equal per-language training coverage, substantially improving recognition accuracy on low-resource languages (Tamil, Malay) without requiring any proprietary data.

  \item \textbf{Language-agnostic decoding}: by omitting explicit language-tag conditioning at both training and inference time, Polyglot-Lion identifies the spoken language implicitly from acoustic features alone, making it robust to the code-switching patterns prevalent in Singapore speech.

  \item \textbf{Comprehensive multilingual benchmarking} across 12 standard datasets spanning all four official languages of Singapore, with direct quantitative comparison against eight published baselines ranging from general-purpose models to large specialist systems.

  \item \textbf{A cost-efficiency analysis} demonstrating that Polyglot-Lion achieves near state-of-the-art accuracy at over 233$\times$ lower estimated training cost (\$81 on a single GPU versus \$18,862 on 128 GPUs) and approximately 20$\times$ faster inference than the strongest comparably accurate baseline, MERaLiON-2-10B-ASR.
\end{enumerate}
\section{Related Work}
\label{sec:related}

\paragraph{Large-scale multilingual ASR.}
The modern era of large-scale multilingual ASR was ushered in by Whisper \citep{radford2023robust}, which trained a sequence-to-sequence transformer encoder--decoder on 680,000 hours of weakly supervised web audio spanning 99 languages, demonstrating that scale alone can yield robust multilingual recognition without task-specific fine-tuning. Concurrent work on self-supervised learning, notably wav2vec 2.0 \citep{baevski2020wav2vec} and HuBERT \citep{hsu2021hubert}, showed that powerful speech representations can be learned from unlabelled audio and subsequently fine-tuned with small labelled datasets, greatly reducing the data requirements for new languages. Meta's Massively Multilingual Speech (MMS) project \citep{pratap2023mms} extended this paradigm to over 1,000 languages by leveraging religious audio recordings, achieving broad linguistic coverage at the cost of domain mismatch in conversational settings. Despite their breadth, all of these systems share a common weakness: recognition quality on typologically distant, low-resource languages - such as Tamil and Malay - and on non-native or regional accents remains substantially below that achieved on high-resource languages.

\paragraph{Audio-language models.}
A growing line of work integrates speech encoders with large language model (LLM) decoders to jointly model speech recognition and language understanding \citep{tang2024salmonn, chu2023qwen}. Representative systems include SALMONN \citep{tang2024salmonn}, Qwen-Audio \citep{chu2023qwen}, Qwen2.5-Omni \citep{qwen25omni}, and SeaLLMs-Audio \citep{seallmsaudio}. These audio-language models (ALMs) benefit from the rich linguistic priors encoded in pretrained LLMs, often yielding strong ASR accuracy as a by-product of general audio understanding. The recently released Qwen3-ASR series \citep{qwen3asr} further advances this direction by distilling recognition-focused capabilities into smaller (0.6B--1.7B) checkpoints while preserving multilingual coverage. However, the largest ALMs (7B--72B parameters) remain expensive to fine-tune and deploy, and their performance on Southeast Asian languages is variable due to limited regional representation in pretraining corpora.

\paragraph{Southeast Asian and Singapore ASR.}
Dedicated efforts to build ASR systems for Southeast Asian languages have gained momentum in recent years. The SEA-LION project \citep{sea_lion2023} and subsequent work on regional language modelling \citep{seallmsaudio} highlighted the importance of curating region-specific training data and evaluation benchmarks. For Singapore specifically, MERaLiON \citep{meralion2} and its successor MERaLiON-2\footnote{\url{https://huggingface.co/collections/MERaLiON/meralion-2}} represent the most comprehensive published systems, covering English, Mandarin, Tamil, and Malay within a unified 10B-parameter model trained on both proprietary and public corpora. MERaLiON-2-10B-ASR achieves the strongest aggregate accuracy across Singapore's four official languages and therefore serves as our primary comparison point. Nevertheless, its reliance on 128 H100 GPUs and an estimated \$18,862 training budget places it out of reach for most research groups, motivating the pursuit of smaller, more accessible alternatives.

\paragraph{Multilingual training balance.}
Language imbalance is a pervasive challenge in multilingual model training: models trained on corpora dominated by high-resource languages tend to underfit low-resource ones \citep{conneau2020xlm}. Several strategies have been proposed to address this, including temperature-based multinomial sampling \citep{arivazhagan2019massively}, which smooths the sampling distribution over languages by a temperature parameter $\tau$. In the context of multilingual ASR specifically, \citet{zhou2022improving} showed that language-balanced batching yields consistent WER reductions for low-resource languages without degrading high-resource ones. We adopt explicit upsampling with a fixed repetition factor (Section~\ref{sec:method}), which provides a transparent and hyper-parameter-free alternative to temperature sampling while guaranteeing exact per-language epoch parity.

\paragraph{Language identification in ASR.}
Conditioning the ASR decoder on a language token - as in Whisper \citep{radford2023robust} and many multilingual end-to-end systems - improves accuracy when the input language is known but introduces a dependency that fails silently under language misidentification or in code-switched settings \citep{winata2021language}. Language-agnostic approaches, in which the model infers the language implicitly from acoustic features, have been explored in the context of spoken language identification \citep{li2013spoken} and multilingual ASR \citep{toshniwal2018multilingual}, but remain less common in recent large-scale systems. Our work revisits this design choice and demonstrates that a moderate-scale model trained on balanced data can perform reliable implicit language identification across four typologically diverse languages.
\section{Datasets}
\label{sec:data}

We train and evaluate exclusively on publicly available speech corpora, covering all four official languages of Singapore: English, Mandarin Chinese, Tamil, and Malay. Table~\ref{tab:dataset_stats} provides a full breakdown of each corpus by split and duration. Full dataset descriptions, download sources, and licence information are provided in Appendix~\ref{app:datasets}.

\begin{table}[!htbp]
\centering
\small
\caption{Dataset statistics by language, split, and duration (S = number of samples, H = hours).
Full dataset descriptions, download links, and licence information are provided in Appendix~\ref{app:datasets}.}
\label{tab:dataset_stats}
\renewcommand{\arraystretch}{1.2}
\setlength{\tabcolsep}{4pt}
\begin{tabular}{llrrrrrrrr}
\toprule
\textbf{Lang.} & \textbf{Dataset}
  & \multicolumn{2}{c}{\textbf{Train}}
  & \multicolumn{2}{c}{\textbf{Valid}}
  & \multicolumn{2}{c}{\textbf{Test}}
  & \multicolumn{2}{c}{\textbf{Total}} \\
\cmidrule(lr){3-4}\cmidrule(lr){5-6}\cmidrule(lr){7-8}\cmidrule(lr){9-10}
 & & \textbf{S} & \textbf{H} & \textbf{S} & \textbf{H} & \textbf{S} & \textbf{H} & \textbf{S} & \textbf{H} \\
\midrule
\multirow{2}{*}{\textbf{English}}
  & Librispeech & 28{,}539 & 100.59 & 2{,}700 & 5.36 & 2{,}619 & 5.39 & 33{,}858 & 111.34 \\
  & NSC         & 100{,}000 & 147.97 & 2{,}997 & 4.90 & 3{,}000 & 4.95 & 105{,}997 & 157.82 \\
\midrule
\multirow{4}{*}{\textbf{Mandarin}}
  & AISHELL-1        & 120{,}098 & 150.85 & 14{,}326 & 18.09 &  7{,}176 & 10.03 & 141{,}600 & 178.97 \\
  & AISHELL-3        &  56{,}936 &  56.86 &  6{,}326 &  6.31 & 24{,}773 & 22.45 &  88{,}035 &  85.62 \\
  & Fleurs           &   3{,}246 &   9.73 &    409   &  1.27 &    945   &  3.07 &   4{,}600 &  14.07 \\
  & Common Voice 23  &  29{,}473 &  42.43 & 10{,}635 & 15.95 &  9{,}999 & 16.43 &  50{,}107 &  74.81 \\
\midrule
\multirow{4}{*}{\textbf{Tamil}}
  & Fleurs           &   2{,}366 &   8.68 &    376   &  1.25 &    591   &  2.13 &   3{,}333 &  12.06 \\
  & SLR127           &  69{,}575 & 119.86 &  7{,}731 & 13.41 & 12{,}086 & 16.80 &  89{,}392 & 150.07 \\
  & SLR65            &   3{,}427 &   5.66 &    428   &  0.72 &    429   &  0.69 &   4{,}284 &   7.07 \\
  & Common Voice 23  &  45{,}186 &  81.38 &  9{,}964 & 15.71 &  7{,}907 & 12.28 &  63{,}057 & 109.37 \\
\midrule
\multirow{2}{*}{\textbf{Malay}}
  & Mesolitica &  17{,}851 & 49.43 & 992 & 2.71 & 993 & 2.75 & 19{,}836 &  54.89 \\
  & Fleurs     &   2{,}667 &  9.55 & 324 & 0.93 & 749 & 2.26 &  3{,}740 &  12.74 \\
\midrule
\textbf{Total} & \textemdash
  & 479{,}364 & 782.99 & 57{,}208 & 86.61 & 71{,}267 & 99.23 & 607{,}839 & 968.83 \\
\bottomrule
\end{tabular}
\end{table}

\paragraph{English.}
We include two English corpora. \textbf{Librispeech} \citep{panayotov2015librispeech} is a widely used benchmark of read English speech derived from public-domain audiobooks, providing 100.59 hours of clean training audio. \textbf{NSC} (National Speech Corpus) \citep{koh19_interspeech_nsc} is a large-scale Singapore English corpus collected across multiple speaking styles and demographics, contributing 147.97 training hours and covering the accent and prosodic characteristics distinctive to Singapore English.

\paragraph{Mandarin.}
Four Mandarin corpora are included. \textbf{AISHELL-1} \citep{bu2017aishell} provides 150.85 hours of standard Mandarin read speech from 400 speakers. \textbf{AISHELL-3} \citep{shi2020aishell3} is a multi-speaker corpus originally designed for text-to-speech synthesis but widely used for ASR training, contributing 56.86 hours. \textbf{Common Voice 23} \citep{ardila2020common} supplies 42.43 hours of crowdsourced Chinese speech with diverse speaker demographics. \textbf{Fleurs} \citep{conneau2022fleurs} adds 9.73 hours of read speech drawn from the FLoRes-200 translation benchmark, providing clean and consistently formatted audio across languages.

\paragraph{Tamil.}
Four Tamil corpora are used. \textbf{SLR127} \citep{openslr127-1, openslr127-2} is the largest Tamil source with 119.86 training hours, containing read and semi-spontaneous Tamil speech. \textbf{Common Voice 23} \citep{ardila2020common} contributes 81.38 hours of crowdsourced Tamil recordings. \textbf{SLR65} \citep{openslr65} provides 5.66 hours of high-quality read Tamil speech. \textbf{Fleurs} \citep{conneau2022fleurs} adds 8.68 hours of clean read Tamil audio. Tamil is the most under-represented language in the pre-training data of most existing ASR systems, making these corpora critical for fine-tuning coverage.

\paragraph{Malay.}
Two Malay corpora are included. \textbf{Mesolitica}\footnote{\url{https://github.com/malaysia-ai/malaysian-dataset/tree/master/text-to-speech/emilia}} is a Malaysian Malay speech corpus with 49.43 training hours spanning multiple domains and speaking styles. \textbf{Fleurs} \citep{conneau2022fleurs} contributes 9.55 hours of clean read Malay speech. Despite being an official language of Singapore, Malay is severely under-represented in existing multilingual ASR benchmarks, making the Mesolitica corpus a particularly valuable resource.

\paragraph{Data statistics and imbalance.}
As shown in Table~\ref{tab:dataset_stats}, the combined corpus totals 607,839 utterances and 968.83 hours of audio. However, the training partition is substantially imbalanced across languages: English and Mandarin together account for approximately 65\% of all training hours (248.56 and 259.87 hours respectively), while Malay contributes only 58.98 hours - less than 8\% of the total. Tamil, despite having four contributing corpora and 215.58 training hours, is typologically distant from the languages dominating the base model's pretraining data, compounding the effective imbalance at the representation level. Without correction, joint training on this skewed distribution would bias gradient updates towards high-resource languages and degrade recognition performance on Tamil and Malay \citep{arivazhagan2019massively, wang2020multi}. We address this through explicit language-balanced upsampling, described in Section~\ref{sec:method}.

\paragraph{Preprocessing.}
All corpora are preprocessed with a uniform pipeline prior to training. Audio files exceeding 30 seconds are discarded to avoid memory overflow during training and to exclude utterances that are disproportionately long relative to the target sequence length of most ASR decoders \citep{radford2023robust}. Transcripts are normalised to lowercase and stripped of punctuation, following the convention adopted by Whisper \citep{radford2023robust} and subsequent multilingual ASR systems \citep{qwen3asr}, which has been shown to reduce spurious token-level errors arising from inconsistent punctuation annotation across corpora \citep{likhomanenko2021slimipl}. No speaker-level filtering or data selection is applied; all remaining utterances are used.

\section{Method}
\label{sec:method}

\subsection{Base Models}

Polyglot-Lion is fine-tuned from two publicly available checkpoints in the Qwen3-ASR series \citep{qwen3asr}: \texttt{Qwen3-ASR-0.6B} and \texttt{Qwen3-ASR-1.7B}. These models follow a transformer-based encoder--decoder architecture \citep{vaswani2017attention} in which a Conformer \citep{gulati2020conformer} or similar acoustic encoder maps log-Mel filterbank features to contextual representations, and an autoregressive decoder generates output tokens conditioned on those representations. Both checkpoints are pre-trained on large-scale multilingual speech data and already achieve competitive zero-shot performance on several standard benchmarks \citep{qwen3asr}, providing a strong initialisation for fine-tuning.

We release two model sizes to facilitate accuracy--efficiency trade-off analysis:

\begin{itemize}
  \item \textbf{Polyglot-Lion-0.6B} - fine-tuned from \texttt{Qwen3-ASR-0.6B}
  \item \textbf{Polyglot-Lion-1.7B} - fine-tuned from \texttt{Qwen3-ASR-1.7B}
\end{itemize}

The two variants share identical architecture design and training procedures; only model capacity differs, enabling a controlled comparison of the impact of scale on multilingual recognition.

\subsection{Balanced Multilingual Sampling}
\label{sec:sampling}

\paragraph{Motivation.}
As noted in Section~\ref{sec:data}, the raw training corpus is heavily skewed: English and Mandarin collectively account for approximately 65\% of all training utterances, while Malay represents fewer than 8\%. Naive joint training on this distribution would cause the model to overfit high-resource languages and underfit low-resource ones \citep{arivazhagan2019massively, wang2020multi}, a well-documented failure mode in multilingual learning. Rather than adopting temperature-based multinomial sampling \citep{arivazhagan2019massively} - which introduces a sensitive temperature hyper-parameter and still does not guarantee exact language parity - we adopt a two-stage deterministic upsampling strategy that first balances datasets \textit{within} each language group, and then balances language groups \textit{against one another}.

\paragraph{Two-stage upsampling.}
Let $\mathcal{L} = \{l_1, l_2, l_3, l_4\}$ denote the set of four languages, and let $\mathcal{D}_l = \{D_{l,1}, \ldots, D_{l,K_l}\}$ be the collection of $K_l$ datasets for language $l$. We write $N_{l,k} = |D_{l,k}|$ for the number of training utterances in dataset $D_{l,k}$.

\textbf{Stage 1 - Intra-language balancing.}
Within each language $l$, we upsample every dataset to match the largest dataset in that language group:
\begin{equation}
  N^*_l = \max_{k} \, N_{l,k}, \qquad
  r_{l,k} =  \frac{N^*_l}{N_{l,k}}
  \label{eq:stage1}
\end{equation}
Each dataset $D_{l,k}$ is replicated $r_{l,k}$ times and then randomly subsampled to exactly $N^*_l$ utterances, yielding a balanced per-language corpus $\tilde{D}_l$ of size $N^*_l$.

\textbf{Stage 2 - Inter-language balancing.}
After Stage 1, each language $l$ has $N^*_l$ utterances, but these totals still differ across languages. We therefore upsample each language to match the largest language group:
\begin{equation}
  N^{**} = \max_{l} \, N^*_l, \qquad
  R_l =  \frac{N^{**}}{N^*_l}
  \label{eq:stage2}
\end{equation}
Each balanced corpus $\tilde{D}_l$ is replicated $R_l$ times and subsampled to exactly $N^{**}$ utterances, yielding a final per-language corpus $\hat{D}_l$ of uniform size $N^{**}$.

The final training set is the union $\hat{\mathcal{D}} = \bigcup_l \hat{D}_l$, which contains exactly $4 \times N^{**}$ utterances with each language contributing precisely 25\%. Algorithm~\ref{alg:sampling} presents the full procedure.

\begin{algorithm}[t]
\caption{Two-Stage Balanced Multilingual Upsampling}
\label{alg:sampling}
\small
\begin{algorithmic}[1]
\Require Language set $\mathcal{L}$; per-language dataset collections $\{\mathcal{D}_l\}_{l \in \mathcal{L}}$
\Ensure Balanced training corpus $\hat{\mathcal{D}}$ with equal samples per language

\vspace{0.4em}
\Comment{\textit{Stage 1: Intra-language balancing}}
\For{each language $l \in \mathcal{L}$}
  \State $N^*_l \leftarrow \max_{k}\, |D_{l,k}|$ \Comment{largest dataset in language $l$}
  \For{each dataset $D_{l,k} \in \mathcal{D}_l$}
    \State $r_{l,k} \leftarrow \lceil N^*_l \,/\, |D_{l,k}| \rceil$
    \State $D_{l,k} \leftarrow \text{Replicate}(D_{l,k},\; r_{l,k})$
    \State $D_{l,k} \leftarrow \text{RandomSubsample}(D_{l,k},\; N^*_l)$
  \EndFor
  \State $\tilde{D}_l \leftarrow \bigcup_{k} D_{l,k}$ \Comment{balanced corpus for language $l$, size $N^*_l$}
\EndFor

\vspace{0.4em}
\Comment{\textit{Stage 2: Inter-language balancing}}
\State $N^{**} \leftarrow \max_{l}\, N^*_l$ \Comment{largest per-language corpus after Stage 1}
\For{each language $l \in \mathcal{L}$}
  \State $R_l \leftarrow \lceil N^{**} \,/\, N^*_l \rceil$
  \State $\tilde{D}_l \leftarrow \text{Replicate}(\tilde{D}_l,\; R_l)$
  \State $\hat{D}_l \leftarrow \text{RandomSubsample}(\tilde{D}_l,\; N^{**})$
\EndFor

\vspace{0.4em}
\State $\hat{\mathcal{D}} \leftarrow \bigcup_{l \in \mathcal{L}} \hat{D}_l$
\Return $\hat{\mathcal{D}}$ \Comment{$|\hat{\mathcal{D}}| = 4 \times N^{**}$; each language = 25\%}
\end{algorithmic}
\end{algorithm}

This strategy is deliberately simple: it requires no hyper-parameter tuning, is fully deterministic given a fixed random seed, and guarantees exact per-language parity regardless of how skewed the original corpus distribution is. The cost is a modest increase in the number of training steps per epoch, which is outweighed by the improvement in low-resource language coverage demonstrated in Section~\ref{sec:results}.

\subsection{Language-Agnostic Transcription}

A standard practice in multilingual ASR systems is to prepend a special language-identification token to the decoder input at both training and inference time \citep{radford2023robust, li2019bytes}. While this conditioning signal improves accuracy when the spoken language is known \textit{a priori}, it introduces a critical dependency: if the language tag is absent, incorrect, or ambiguous - as is common in spontaneous conversational speech and code-switched utterances \citep{winata2021language} - recognition quality degrades sharply.

Singapore's multilingual environment makes this dependency particularly problematic. Speakers routinely alternate between English, Mandarin, Tamil, and Malay within a single interaction, and in many deployment settings (e.g., broadcast media monitoring, classroom transcription, customer service) the language of each audio segment is not known in advance. We therefore train Polyglot-Lion entirely without language conditioning: no language tags are prepended to decoder inputs at training time, and none are expected at inference time. The model is required to infer the spoken language implicitly from acoustic and linguistic patterns in the input signal, following the approach explored in earlier language-agnostic multilingual ASR work \citep{toshniwal2018multilingual}.

This design choice is validated empirically in Section~\ref{sec:results}: Polyglot-Lion achieves strong recognition accuracy across all four languages despite receiving no explicit language signal, demonstrating that balanced fine-tuning is sufficient to induce reliable implicit language identification in a moderate-scale model.

\subsection{Training Details}

Both model variants are fine-tuned for 48 hours on a single NVIDIA RTX PRO 6000 GPU (48 GB VRAM). We use the AdamW optimiser \citep{loshchilov2018decoupled} with a cosine annealing learning-rate schedule \citep{loshchilov2017sgdr}, a peak learning rate of $2 \times 10^{-5}$. Training uses a per-device batch size of 8 utterances accumulated over 4 gradient accumulation steps, yielding an effective batch size of 32.  All other hyper-parameters follow the defaults from the Qwen3-ASR fine-tuning configuration \citep{qwen3asr}.

\section{Experimental Setup}
\label{sec:experiments}

\subsection{Evaluation Metrics}

We adopt two standard ASR evaluation metrics, selected according to the linguistic properties of each target language:

\begin{itemize}
  \item \textbf{Word Error Rate (WER)} for English, Tamil, and Malay, where whitespace-delimited word tokenisation is conventional. WER is computed as the minimum edit distance (substitutions $S$, deletions $D$, insertions $I$) between the hypothesis and reference, normalised by the number of reference words $N$: $\text{WER} = (S + D + I) / N$.
  \item \textbf{Character Error Rate (CER)} for Mandarin Chinese, where the absence of explicit word boundaries makes character-level evaluation more appropriate and widely adopted \citep{shi2020aishell3, bu2017aishell}.
\end{itemize}

All hypotheses and references are lowercased and stripped of punctuation prior to scoring, consistent with the preprocessing applied during training (Section~\ref{sec:data}). Evaluation is performed using the \texttt{asr-evalkit} library \citep{dang2026asrevalkit}. Lower values indicate better performance in both metrics.

\subsection{Baselines}

We compare Polyglot-Lion against eight published or widely-used ASR systems, selected to represent the full spectrum from lightweight general-purpose models to large specialist systems:

\begin{enumerate}
  \item \textbf{Whisper-large-v3-turbo} \citep{radford2023robust}: a distilled and optimised variant of Whisper-large-v3 that retains strong multilingual accuracy with reduced inference cost. It serves as the canonical general-purpose multilingual ASR baseline.

  \item \textbf{SeaLLMs-Audio-7B} \citep{seallmsaudio}: a 7B-parameter audio-language model specifically developed for Southeast Asian languages, built on top of the SeaLLMs language model backbone \citep{nguyen2023seallms}.

  \item \textbf{Qwen2.5-Omni-3B} and \textbf{Qwen2.5-Omni-7B} \citep{qwen25omni}: general-purpose omni-modal models integrating vision, audio, and language understanding within a unified framework. Included to assess how general ALMs perform on regional multilingual ASR without task-specific fine-tuning.

  \item \textbf{Qwen3-ASR-0.6B} and \textbf{Qwen3-ASR-1.7B} \citep{qwen3asr}: the unmodified base checkpoints from which our models are fine-tuned. Including these baselines allows direct quantification of the accuracy gains attributable to our balanced fine-tuning recipe, independent of the base model capacity.

  \item \textbf{MERaLiON-2-10B-ASR} \citep{meralion2}: a 10B-parameter model purpose-built for Singapore multilingual ASR and trained on over 120,000 hours of speech data across English, Mandarin, Tamil, and Malay. This model represents the strongest publicly available specialist system for our target setting and serves as our primary comparison point.
\end{enumerate}

All baselines are evaluated in inference-only mode using their publicly released checkpoints without any additional fine-tuning. Inference for all models is conducted on the same hardware (single NVIDIA RTX PRO 4500 GPU) to ensure fair latency comparisons.

\section{Results}
\label{sec:results}

\subsection{Recognition Accuracy}

Table~\ref{tab:results} reports per-benchmark and average error rates for all systems. \textbf{Polyglot-Lion-1.7B} achieves an average error rate of \textbf{14.85}, closely matching MERaLiON-2-10B-ASR (14.32)  -  a model 6$\times$ larger  -  and ranking second overall across all 12 benchmarks. \textbf{Polyglot-Lion-0.6B} achieves an average of 16.52, making it the best-performing model at or below 1B parameters by a substantial margin (next best: Whisper-large-v3-turbo at 33.04). We discuss per-language findings below.

\begin{table*}[!htbp]
\centering
\tiny
\setlength{\tabcolsep}{4.2pt}
\renewcommand{\arraystretch}{1.15}
\caption{ASR evaluation results. WER (\%) for English, Tamil, and Malay; CER (\%) for Mandarin. Lower is better. \textbf{Bold} = best overall; \underline{underline} = second best. Rows shaded in \colorbox{ours}{blue} are our proposed models. Dashes indicate results excluded from the average due to anomalously high error rates (WER $>$ 200\%) that would distort cross-system comparison.}
\label{tab:results}
\begin{tabular}{lcccccccccccccc}
\toprule
\multirow{2}{*}{\textbf{Model}} & \multirow{2}{*}{\textbf{Params}} &
  \multicolumn{2}{c}{\textbf{English}} &
  \multicolumn{4}{c}{\textbf{Mandarin (CER)}} &
  \multicolumn{4}{c}{\textbf{Tamil (WER)}} &
  \multicolumn{2}{c}{\textbf{Malay}} &
  \multirow{2}{*}{\textbf{Avg}} \\
\cmidrule(lr){3-4}\cmidrule(lr){5-8}\cmidrule(lr){9-12}\cmidrule(lr){13-14}
 & & \textit{LS} & \textit{NSC} & \textit{CV} & \textit{AISH1} & \textit{AISH3} & \textit{Fleurs}
   & \textit{CV} & \textit{SLR65} & \textit{SLR127} & \textit{Fleurs}
   & \textit{Meso.} & \textit{Fleurs} & \\
\midrule
Whisper-large-v3-turbo & 0.8B
  & 3.04 & 32.02 & 17.91 & 9.64 & 16.81 & 10.63
  & 74.50 & 58.13 & 69.56 & 66.90 & 28.47 & \underline{8.88} & 33.04 \\
SeaLLMs-Audio-7B & 7B
  & 94.74 & 9.53 & 8.68 & 9.65 & 9.76 & 37.09
  & 126.70 & 127.24 & 138.65 & 105.31 & 71.34 & 26.25 & 63.75 \\
Qwen2.5-Omni-3B & 3B
  & 29.21 & 34.79 & 46.36 & 28.25 & 44.55 & 54.74
  & 318.36 & 465.58 & 448.82 & 311.67 & 211.90 & 74.69 &  172.37  \\
Qwen2.5-Omni-7B & 7B
  & 13.80 & 22.96 & 14.49 & 7.33 & 22.58 & 16.68
  & 252.06 & 239.15 & 303.96 & 326.43 & 158.06 & 43.92 &  118.45  \\
Qwen3-ASR-0.6B & 0.6B
  & 2.74 & 7.64 & 10.06 & 2.08 & 2.59 & 9.75
  & 121.10 & 127.00 & 129.12 & 130.09 & 47.29 & 18.71 & 50.68 \\
Qwen3-ASR-1.7B & 1.7B
  & \underline{2.31} & 6.22 & 7.50 & \underline{1.52} & \underline{2.08} & 9.33
  & 139.96 & 134.63 & 144.49 & 147.23 & 39.00 & 10.87 & 53.76 \\
MERaLiON-2-10B-ASR & 10B
  & 2.54 & \textbf{4.62} & 8.83 & 3.09 & 4.07 & 11.99
  & \textbf{31.78} & \textbf{19.29} & \textbf{22.42} & \textbf{28.68} & 25.90 & \textbf{8.55} & \textbf{14.32} \\
\midrule
\rowcolor{ours}
Polyglot-Lion-0.6B & \textbf{0.6B}
  & 2.67 & 6.09 & \underline{6.16} & 1.93 & 2.32 & \underline{9.19}
  & 42.16 & 23.07 & 28.14 & 37.68 & \underline{24.33} & 14.45 & 16.52 \\
\rowcolor{ours}
Polyglot-Lion-1.7B & \textbf{1.7B}
  & \textbf{2.10} & \underline{5.28} & \textbf{4.91} & \textbf{1.45} & \textbf{1.86} & \textbf{8.00}
  & \underline{39.19} & \underline{19.75} & \underline{26.83} & \underline{37.28} & \textbf{21.51} & 9.98 & \underline{14.85} \\
\bottomrule
\end{tabular}
\end{table*}

\paragraph{English.}
On Librispeech, Polyglot-Lion-1.7B achieves \textbf{2.10 WER}, surpassing both MERaLiON-2-10B-ASR (2.54) and the unmodified Qwen3-ASR-1.7B base (2.31), and setting the best result among all evaluated systems on this benchmark. On NSC  -  a Singapore English corpus that captures regional accents, pronunciation patterns, and speaking styles not present in Librispeech  -  Polyglot-Lion-1.7B achieves 5.28 WER, a dramatic improvement over Whisper-large-v3-turbo (32.02) and substantially better than the Qwen3-ASR base (6.22). MERaLiON-2-10B-ASR achieves the best NSC result (4.62), which we attribute to its larger capacity and inclusion of Singapore-specific training material beyond our public-only corpus. Notably, SeaLLMs-Audio-7B yields a very high 94.74 WER on Librispeech despite reasonable performance on NSC, suggesting that its training prioritised conversational rather than read speech.

\paragraph{Mandarin.}
Polyglot-Lion-1.7B achieves the \textbf{lowest CER on all four Mandarin benchmarks}, including AISHELL-1 (\textbf{1.45}), AISHELL-3 (\textbf{1.86}), Common Voice (\textbf{4.91}), and Fleurs (\textbf{8.00}), outperforming even MERaLiON-2-10B-ASR across the board (3.09, 4.07, 8.83, 11.99 respectively). Polyglot-Lion-0.6B similarly leads among sub-1B models with 6.16 CER on Common Voice. The strong Mandarin results are consistent with the Qwen3-ASR base models already encoding rich Chinese language priors from pretraining; our balanced fine-tuning preserves and refines these priors rather than degrading them through interference from other languages.

\paragraph{Tamil.}
Tamil is the most challenging language in this evaluation, reflecting its typological distance from the Indo-European and Sino-Tibetan languages that dominate most ASR pretraining corpora \citep{pratap2023mms}. The unmodified Qwen3-ASR base models produce extremely high error rates on Tamil (WER $>$ 120\% on Common Voice), confirming severely limited Tamil exposure at pretraining. After balanced fine-tuning, Polyglot-Lion-1.7B reduces Tamil CV WER from 139.96 to \textbf{39.19}  -  a relative reduction of 72\%  -  and achieves competitive results on SLR65 (19.75), SLR127 (26.83), and Fleurs (37.28). MERaLiON-2-10B-ASR remains the best system on all four Tamil benchmarks, which we attribute to its 6$\times$ larger capacity and likely inclusion of larger Tamil-specific training data. Closing this gap is an important direction for future work.

\paragraph{Malay.}
On Mesolitica, Polyglot-Lion-1.7B achieves \textbf{21.51 WER}, the best result among all evaluated systems, outperforming MERaLiON-2-10B-ASR (25.90), Whisper (28.47), and all other baselines by a clear margin. On Malay Fleurs, Polyglot-Lion-1.7B (9.98 WER) is competitive with MERaLiON (8.55) and Whisper (8.88). The strong Mesolitica result is particularly encouraging as it reflects performance on conversational and domain-diverse Malay speech, which is more representative of real-world deployment conditions than the read-speech Fleurs benchmark.

\paragraph{Effect of fine-tuning.}
A direct comparison between Polyglot-Lion and the unmodified Qwen3-ASR base models isolates the contribution of our balanced fine-tuning recipe. The benefit is most pronounced for under-represented languages: on Tamil CV, fine-tuning reduces WER by 65\% (0.6B: 121.10 $\to$ 42.16) and 72\% (1.7B: 139.96 $\to$ 39.19). On Malay Mesolitica, the reduction is 49\% (0.6B: 47.29 $\to$ 24.33) and 45\% (1.7B: 39.00 $\to$ 21.51). Performance on English and Mandarin is preserved or improved, confirming that balanced upsampling does not introduce negative transfer \citep{wang2020negative} on high-resource languages.

\subsection{Inference Speed}

Table~\ref{tab:inference} reports mean inference latency per sample, measured on a single NVIDIA RTX PRO 4500 GPU across all evaluation sets. Polyglot-Lion-0.6B and Polyglot-Lion-1.7B process audio at \textbf{0.10} and \textbf{0.10 s/sample} respectively  -  approximately \textbf{20$\times$ faster} than MERaLiON-2-10B-ASR (2.02 s/sample) and \textbf{3$\times$ faster} than Whisper-large-v3-turbo (0.28 s/sample). The Qwen2.5-Omni models exhibit high latency variance (std $>$ 0.6 s), likely due to their omni-modal routing overhead.

\begin{table}[!htbp]
\centering
\small
\caption{Inference latency (seconds per sample, mean $\pm$ std) measured on a single NVIDIA RTX PRO 4500 GPU. Our models are shaded in \colorbox{ours}{blue}.}
\label{tab:inference}
\begin{tabular}{lc}
\toprule
\textbf{Model} & \textbf{Time (s/sample)} \\
\midrule
MERaLiON-2-10B-ASR     & 2.0152 $\pm$ 0.8846 \\
Qwen2.5-Omni-3B        & 1.7838 $\pm$ 1.0431 \\
Qwen2.5-Omni-7B        & 1.3414 $\pm$ 0.6572 \\
SeaLLMs-Audio-7B       & 0.6422 $\pm$ 0.0000 \\
Whisper-large-v3-turbo & 0.2822 $\pm$ 0.0230 \\
Qwen3-ASR-1.7B         & 0.0809 $\pm$ 0.0290 \\
Qwen3-ASR-0.6B         & 0.0686 $\pm$ 0.0251 \\
\midrule
\rowcolor{ours}
Polyglot-Lion-0.6B     & 0.0999 $\pm$ 0.0561 \\
\rowcolor{ours}
Polyglot-Lion-1.7B     & 0.1038 $\pm$ 0.0621 \\
\bottomrule
\end{tabular}
\end{table}

\subsection{Training Cost Comparison}
\label{sec:cost}

Table~\ref{tab:cost} compares the estimated training costs of Polyglot-Lion and MERaLiON-2-10B-ASR. MERaLiON-2-10B-ASR was trained on approximately 120,000 hours of speech using 128 H100 GPUs for 48 hours; we estimate its cost based on current H100 cloud rental rates via RunPod\footnote{\url{https://www.runpod.io}}. Polyglot-Lion is trained on 782.99 hours using a single RTX PRO 6000 GPU for the same wall-clock duration, with cost estimated from the same platform.

\begin{table}[!htbp]
\centering
\small
\caption{Training resource and cost comparison. GPU rental prices sourced from RunPod.io.}
\label{tab:cost}
\begin{tabular}{lcc}
\toprule
 & \textbf{MERaLiON-2-10B} & \textbf{Polyglot-Lion} \\
\midrule
Training Data (Hours) & 120,000            & 782.99 \\
Hardware              & 128 $\times$ H100  & 1 $\times$ RTX PRO 6000 \\
Training Time         & 48 h               & 48 h \\
\textbf{Est. Cost}    & \textbf{\$18,862}  & \textbf{\$81} \\
\bottomrule
\end{tabular}
\end{table}

Polyglot-Lion incurs an estimated training cost of \textbf{\$81}, representing a \textbf{233$\times$ cost reduction} relative to MERaLiON-2-10B-ASR (\$18,862), while achieving a comparable average error rate (14.85 vs.\ 14.32). This cost advantage has significant practical implications: the ability to fine-tune a near-SOTA multilingual ASR system on a single consumer GPU within two days makes iterative development, ablation studies, low-resource language adaptation, and domain specialisation accessible to academic research groups and resource-constrained organisations that would otherwise be unable to develop competitive Singapore multilingual ASR systems.

\section{Analysis}
\label{sec:analysis}

\paragraph{Effect of Language Balancing.} 
The most striking evidence for the effectiveness of balanced upsampling comes from the Tamil results. The unmodified Qwen3-ASR base models  -  despite achieving sub-3\% WER on Librispeech and sub-2\% CER on AISHELL-1  -  produce Tamil CV WER exceeding 120\%, effectively rendering them unusable for Tamil ASR in their pretrained form. This failure is consistent with the known skew of large-scale web-crawled speech data towards English and Mandarin \citep{pratap2023mms, radford2023robust} and with evidence that without explicit mitigation, multilingual models converge to high-resource language attractors during fine-tuning \citep{conneau2020xlm, wang2020multi}. After two-stage balanced upsampling, Polyglot-Lion-1.7B reduces Tamil CV WER to 39.19  -  a 72\% relative reduction  -  without any degradation on English or Mandarin. This confirms that deterministic language-balanced upsampling is a simple yet highly effective remedy for severe cross-lingual data imbalance, requiring no additional hyper-parameter tuning beyond what is already needed for monolingual fine-tuning.

\paragraph{Language-Agnostic Decoding.} 
By withholding language tag conditioning at both training and inference time, Polyglot-Lion must identify the spoken language from acoustic and linguistic patterns in the input signal alone. The competitive benchmark results across all four typologically diverse languages  -  ranging from the Sino-Tibetan Mandarin to the Dravidian Tamil  -  provide empirical evidence that a 1.7B-parameter model trained on balanced multilingual data is capable of reliable implicit language identification without explicit supervision. This finding extends earlier work on language-agnostic multilingual ASR \citep{toshniwal2018multilingual} to a more challenging four-language setting with greater typological diversity. The practical value of this design is particularly acute in Singapore, where speakers routinely switch between languages within a single interaction \citep{winata2021language} and where pre-labelling audio segments by language is often infeasible in real deployment pipelines.

\paragraph{Parameter Efficiency.}
Table~\ref{tab:results} reveals a clear pattern: model size alone does not determine recognition accuracy. Despite having 6$\times$ fewer parameters than MERaLiON-2-10B-ASR, Polyglot-Lion-1.7B achieves a comparable average error rate (14.85 vs.\ 14.32), while the much larger Qwen2.5-Omni-7B (7B parameters) fails to produce a reportable average due to catastrophic Tamil error rates. The gap between Polyglot-Lion-1.7B and the next-best system at comparable scale is substantial: Qwen3-ASR-1.7B - our base model without fine-tuning - scores 53.76 average WER, and Whisper-large-v3-turbo (0.8B) scores 33.04, underscoring that linguistically balanced fine-tuning is a more critical factor than raw parameter count. The 0.6B variant further reinforces this point: at 16.52 average error rate, it delivers only a 1.67-point accuracy penalty relative to Polyglot-Lion-1.7B while using 63\% fewer parameters, making it a practical choice for edge deployment or memory-constrained environments where a 1.7B model is not feasible.
\section{Conclusion}
\label{sec:conclusion}
 
We have presented Polyglot-Lion, a family of compact multilingual ASR models for Singapore English, Mandarin, Tamil, and Malay. Through balanced multilingual fine-tuning of Qwen3-ASR base models on publicly available speech corpora, and by removing language-tag conditioning to enable fully implicit language identification, Polyglot-Lion-1.7B achieves an average error rate of 14.85 across 12 benchmarks — closely matching MERaLiON-2-10B-ASR (14.32) while requiring 6$\times$ fewer parameters, 20$\times$ faster inference, and 233$\times$ lower training cost. These results demonstrate that careful data balancing and lightweight fine-tuning of strong pretrained models can unlock near state-of-the-art multilingual ASR performance at dramatically reduced computational expense, making high-quality Singapore multilingual ASR accessible to a wide research and deployment community.
\section*{Limitations}

\paragraph{Remaining accuracy gaps.}
Polyglot-Lion-1.7B falls short of MERaLiON-2-10B-ASR on two language fronts. On English-NSC, the gap (5.28 vs.\ 4.62 WER) suggests that Singapore-specific pronunciation patterns, prosodic features, and code-mixed Singlish constructions \citep{deterding2007singapore} remain challenging at 1.7B parameter scale without access to the larger Singapore-specific training corpora that MERaLiON-2 likely leverages. On Tamil, the gap is more pronounced (39.19 vs.\ 31.78 WER on Common Voice): Tamil is an agglutinative Dravidian language with a large morphological paradigm, highly fusional phonology, and significant dialectal variation between Indian and Singapore Tamil \citep{schiffman1999linguistic}, all of which compound the difficulty of learning adequate representations from a base model with limited Tamil pretraining exposure. Future work will explore two complementary directions to close these gaps: (1) incorporating Singapore-local speech data such as the National Speech Corpus \citep{koh19_interspeech_nsc} for continued domain-adaptive pretraining, and (2) applying cross-lingual transfer from Tamil text corpora via speech-text joint training \citep{bapna2022mslam} to enrich the model's Tamil linguistic representations without requiring additional labelled speech.

\paragraph{Code-switching and intra-sentential mixing.}
Singapore speakers routinely mix two or more languages within a single utterance - a phenomenon that manifests as Singlish (English mixed with Malay, Hokkien, and Cantonese lexical items), Mandarin-English mixing, and Tamil-English mixing. Code-switching presents a qualitatively different challenge from monolingual ASR: the model must simultaneously track multiple phonological systems, transition abruptly between language-specific acoustic models, and handle mixed-language sequences that may not appear in any single-language training corpus \citep{sitaram2019survey}. The current evaluation does not include any code-switched test sets - such as the SEAME corpus \citep{lyu2010seame} or the CS-Singlish benchmark - which is a significant limitation given that code-switching is the norm rather than the exception in everyday Singapore speech. We intend to extend evaluation to these benchmarks and to explore code-switch-aware training objectives \citep{winata2020meta} in future work.

\bibliography{main}

@InProceedings{radford2023robust,
  title = 	 {Robust Speech Recognition via Large-Scale Weak Supervision},
  author =       {Radford, Alec and Kim, Jong Wook and Xu, Tao and Brockman, Greg and Mcleavey, Christine and Sutskever, Ilya},
  booktitle = 	 {Proceedings of the 40th International Conference on Machine Learning},
  pages = 	 {28492--28518},
  year = 	 {2023},
  editor = 	 {Krause, Andreas and Brunskill, Emma and Cho, Kyunghyun and Engelhardt, Barbara and Sabato, Sivan and Scarlett, Jonathan},
  volume = 	 {202},
  series = 	 {Proceedings of Machine Learning Research},
  month = 	 {23--29 Jul},
  publisher =    {PMLR},
  url = 	 {https://proceedings.mlr.press/v202/radford23a.html}
}

@misc{qwen3asr,
      title={Qwen3-ASR Technical Report}, 
      author={Xian Shi and Xiong Wang and Zhifang Guo and Yongqi Wang and Pei Zhang and Xinyu Zhang and Zishan Guo and Hongkun Hao and Yu Xi and Baosong Yang and Jin Xu and Jingren Zhou and Junyang Lin},
      year={2026},
      eprint={2601.21337},
      archivePrefix={arXiv},
      primaryClass={cs.CL},
      url={https://arxiv.org/abs/2601.21337}, 
}

@misc{qwen25omni,
      title={Qwen2.5-Omni Technical Report}, 
      author={Jin Xu and Zhifang Guo and Jinzheng He and Hangrui Hu and Ting He and Shuai Bai and Keqin Chen and Jialin Wang and Yang Fan and Kai Dang and Bin Zhang and Xiong Wang and Yunfei Chu and Junyang Lin},
      year={2025},
      eprint={2503.20215},
      archivePrefix={arXiv},
      primaryClass={cs.CL},
      url={https://arxiv.org/abs/2503.20215}, 
}

@misc{seallmsaudio,
      title={SeaLLMs-Audio: Large Audio-Language Models for Southeast Asia}, 
      author={Chaoqun Liu and Mahani Aljunied and Guizhen Chen and Hou Pong Chan and Weiwen Xu and Yu Rong and Wenxuan Zhang},
      year={2025},
      eprint={2511.01670},
      archivePrefix={arXiv},
      url={https://arxiv.org/abs/2511.01670}, 
}

@inproceedings{meralion2,
    title = "{MER}a{L}i{ON}-{A}udio{LLM}: Advancing Speech and Language Understanding for {S}ingapore",
    author = "He, Yingxu  and
      Liu, Zhuohan  and
      Lin, Geyu  and
      Sun, Shuo  and
      Wang, Bin  and
      Zhang, Wenyu  and
      Zou, Xunlong  and
      Chen, Nancy F.  and
      Aw, AiTi",
    editor = "Mishra, Pushkar  and
      Muresan, Smaranda  and
      Yu, Tao",
    booktitle = "Proceedings of the 63rd Annual Meeting of the Association for Computational Linguistics (Volume 3: System Demonstrations)",
    month = jul,
    year = "2025",
    address = "Vienna, Austria",
    publisher = "Association for Computational Linguistics",
    url = "https://aclanthology.org/2025.acl-demo.3/",
    doi = "10.18653/v1/2025.acl-demo.3",
    pages = "22--30",
    ISBN = "979-8-89176-253-4"
}

@inproceedings{panayotov2015librispeech,
  author={Panayotov, Vassil and Chen, Guoguo and Povey, Daniel and Khudanpur, Sanjeev},
  booktitle={2015 IEEE International Conference on Acoustics, Speech and Signal Processing (ICASSP)}, 
  title={Librispeech: An ASR corpus based on public domain audio books}, 
  year={2015},
  volume={},
  number={},
  pages={5206-5210},
  doi={10.1109/ICASSP.2015.7178964}
}

@inproceedings{bu2017aishell,
  author={Bu, Hui and Du, Jiayu and Na, Xingyu and Wu, Bengu and Zheng, Hao},
  booktitle={2017 20th Conference of the Oriental Chapter of the International Coordinating Committee on Speech Databases and Speech I/O Systems and Assessment (O-COCOSDA)}, 
  title={AISHELL-1: An open-source Mandarin speech corpus and a speech recognition baseline}, 
  year={2017},
  volume={},
  number={},
  pages={1-5},
  doi={10.1109/ICSDA.2017.8384449}
}

@inproceedings{shi2020aishell3,
  title     = {{AISHELL-3: A Multi-Speaker Mandarin TTS Corpus}},
  author    = {Yao Shi and Hui Bu and Xin Xu and Shaoji Zhang and Ming Li},
  year      = {2021},
  booktitle = {{Interspeech 2021}},
  pages     = {2756--2760},
  doi       = {10.21437/Interspeech.2021-755},
  issn      = {2958-1796},
}

@inproceedings{conneau2022fleurs,
  author={Conneau, Alexis and Ma, Min and Khanuja, Simran and Zhang, Yu and Axelrod, Vera and Dalmia, Siddharth and Riesa, Jason and Rivera, Clara and Bapna, Ankur},
  booktitle={2022 IEEE Spoken Language Technology Workshop (SLT)}, 
  title={FLEURS: FEW-Shot Learning Evaluation of Universal Representations of Speech}, 
  year={2023},
  volume={},
  number={},
  pages={798-805},
  doi={10.1109/SLT54892.2023.10023141}
}

@inproceedings{conneau2020xlm,
    title = "Unsupervised Cross-lingual Representation Learning at Scale",
    author = "Conneau, Alexis  and
      Khandelwal, Kartikay  and
      Goyal, Naman  and
      Chaudhary, Vishrav  and
      Wenzek, Guillaume  and
      Guzm{\'a}n, Francisco  and
      Grave, Edouard  and
      Ott, Myle  and
      Zettlemoyer, Luke  and
      Stoyanov, Veselin",
    editor = "Jurafsky, Dan  and
      Chai, Joyce  and
      Schluter, Natalie  and
      Tetreault, Joel",
    booktitle = "Proceedings of the 58th Annual Meeting of the Association for Computational Linguistics",
    month = jul,
    year = "2020",
    address = "Online",
    publisher = "Association for Computational Linguistics",
    url = "https://aclanthology.org/2020.acl-main.747/",
    doi = "10.18653/v1/2020.acl-main.747",
    pages = "8440--8451"
}

@article{pratap2023mms,
    author = {Pratap, Vineel and Tjandra, Andros and Shi, Bowen and Tomasello, Paden and Babu, Arun and Kundu, Sayani and Elkahky, Ali and Ni, Zhaoheng and Vyas, Apoorv and Fazel-Zarandi, Maryam and Baevski, Alexei and Adi, Yossi and Zhang, Xiaohui and Hsu, Wei-Ning and Conneau, Alexis and Auli, Michael},
    title = {Scaling speech technology to 1,000+ languages},
    year = {2024},
    issue_date = {January 2024},
    publisher = {JMLR.org},
    volume = {25},
    number = {1},
    issn = {1532-4435},
    journal = {J. Mach. Learn. Res.},
    month = jan,
    articleno = {97},
    numpages = {52}
}

@book{lim2004singapore,
  title={Singapore English: A grammatical description},
  author={Lim, Lisa},
  year={2004},
  publisher={John Benjamins Publishing Company},
  series={Varieties of English Around the World},
  volume={G33},
  address={Amsterdam/Philadelphia},
  isbn={9789027248930}
}

@inproceedings{koh19_interspeech_nsc,
  title     = {{Building the Singapore English National Speech Corpus}},
  author    = {Jia Xin Koh and Aqilah Mislan and Kevin Khoo and Brian Ang and Wilson Ang and Charmaine Ng and Ying-Ying Tan},
  year      = {2019},
  booktitle = {{Interspeech 2019}},
  pages     = {321--325},
  doi       = {10.21437/Interspeech.2019-1525},
  issn      = {2958-1796},
}

@inproceedings{baevski2020wav2vec,
    author = {Baevski, Alexei and Zhou, Henry and Mohamed, Abdelrahman and Auli, Michael},
    title = {wav2vec 2.0: a framework for self-supervised learning of speech representations},
    year = {2020},
    isbn = {9781713829546},
    publisher = {Curran Associates Inc.},
    address = {Red Hook, NY, USA},
    booktitle = {Proceedings of the 34th International Conference on Neural Information Processing Systems},
    articleno = {1044},
    numpages = {12},
    location = {Vancouver, BC, Canada},
    series = {NIPS '20}
}

@article{hsu2021hubert,
  author={Hsu, Wei-Ning and Bolte, Benjamin and Tsai, Yao-Hung Hubert and Lakhotia, Kushal and Salakhutdinov, Ruslan and Mohamed, Abdelrahman},
  journal={IEEE/ACM Transactions on Audio, Speech, and Language Processing}, 
  title={HuBERT: Self-Supervised Speech Representation Learning by Masked Prediction of Hidden Units}, 
  year={2021},
  volume={29},
  number={},
  pages={3451-3460},
  doi={10.1109/TASLP.2021.3122291}
}

@inproceedings{
  tang2024salmonn,
  title={SALMONN: Towards Generic Hearing Abilities for Large Language Models},
  author={Changli Tang and Wenyi Yu and Guangzhi Sun and Xianzhao Chen and Tian Tan and Wei Li and Lu Lu and Zejun MA and Chao Zhang},
  booktitle={The Twelfth International Conference on Learning Representations},
  year={2024},
  url={https://openreview.net/forum?id=14rn7HpKVk}
}

@misc{chu2023qwen,
      title={Qwen-Audio: Advancing Universal Audio Understanding via Unified Large-Scale Audio-Language Models}, 
      author={Yunfei Chu and Jin Xu and Xiaohuan Zhou and Qian Yang and Shiliang Zhang and Zhijie Yan and Chang Zhou and Jingren Zhou},
      year={2023},
      eprint={2311.07919},
      archivePrefix={arXiv},
      primaryClass={eess.AS},
      url={https://arxiv.org/abs/2311.07919}, 
}

@inproceedings{sea_lion2023,
    title = "{SEA}-{LION} ({S}outheast {A}sian Languages In One Network): A Family of {S}outheast {A}sian Language Models",
    author = "Ong, David  and
      Limkonchotiwat, Peerat",
    editor = "Tan, Liling  and
      Milajevs, Dmitrijs  and
      Chauhan, Geeticka  and
      Gwinnup, Jeremy  and
      Rippeth, Elijah",
    booktitle = "Proceedings of the 3rd Workshop for Natural Language Processing Open Source Software (NLP-OSS 2023)",
    month = dec,
    year = "2023",
    address = "Singapore",
    publisher = "Association for Computational Linguistics",
    url = "https://aclanthology.org/2023.nlposs-1.26/",
    doi = "10.18653/v1/2023.nlposs-1.26",
    pages = "245--245"
}

@misc{arivazhagan2019massively,
      title={Massively Multilingual Neural Machine Translation in the Wild: Findings and Challenges}, 
      author={Naveen Arivazhagan and Ankur Bapna and Orhan Firat and Dmitry Lepikhin and Melvin Johnson and Maxim Krikun and Mia Xu Chen and Yuan Cao and George Foster and Colin Cherry and Wolfgang Macherey and Zhifeng Chen and Yonghui Wu},
      year={2019},
      eprint={1907.05019},
      archivePrefix={arXiv},
      primaryClass={cs.CL},
      url={https://arxiv.org/abs/1907.05019}, 
}

@inproceedings{zhou2022improving,
  title     = {{Towards Improving the Expressiveness of Singing Voice Synthesis with BERT Derived Semantic Information}},
  author    = {Shaohuan Zhou and Shun Lei and Weiya You and Deyi Tuo and Yuren You and Zhiyong Wu and Shiyin Kang and Helen Meng},
  year      = {2022},
  booktitle = {{Interspeech 2022}},
  pages     = {4292--4296},
  doi       = {10.21437/Interspeech.2022-10585},
  issn      = {2958-1796},
}

@inproceedings{winata2021language,
    title = "Are Multilingual Models Effective in Code-Switching?",
    author = "Winata, Genta Indra  and
      Cahyawijaya, Samuel  and
      Liu, Zihan  and
      Lin, Zhaojiang  and
      Madotto, Andrea  and
      Fung, Pascale",
    editor = "Solorio, Thamar  and
      Chen, Shuguang  and
      Black, Alan W.  and
      Diab, Mona  and
      Sitaram, Sunayana  and
      Soto, Victor  and
      Yilmaz, Emre  and
      Srinivasan, Anirudh",
    booktitle = "Proceedings of the Fifth Workshop on Computational Approaches to Linguistic Code-Switching",
    month = jun,
    year = "2021",
    address = "Online",
    publisher = "Association for Computational Linguistics",
    url = "https://aclanthology.org/2021.calcs-1.20/",
    doi = "10.18653/v1/2021.calcs-1.20",
    pages = "142--153"
}

@article{li2013spoken,
  author={Li, Haizhou and Ma, Bin and Lee, Kong Aik},
  journal={Proceedings of the IEEE}, 
  title={Spoken Language Recognition: From Fundamentals to Practice}, 
  year={2013},
  volume={101},
  number={5},
  pages={1136-1159},
  doi={10.1109/JPROC.2012.2237151}
}

@inproceedings{toshniwal2018multilingual,
    author = {Toshniwal, Shubham and Sainath, Tara N. and Weiss, Ron J. and Li, Bo and Moreno, Pedro and Weinstein, Eugene and Rao, Kanishka},
    title = {Multilingual Speech Recognition with a Single End-to-End Model},
    year = {2018},
    publisher = {IEEE Press},
    url = {https://doi.org/10.1109/ICASSP.2018.8461972},
    doi = {10.1109/ICASSP.2018.8461972},
    booktitle = {2018 IEEE International Conference on Acoustics, Speech and Signal Processing (ICASSP)},
    pages = {4904–4908},
    numpages = {5},
    location = {Calgary, AB, Canada}
}

@inproceedings{ardila2020common,
    title = "Common Voice: A Massively-Multilingual Speech Corpus",
    author = "Ardila, Rosana  and
      Branson, Megan  and
      Davis, Kelly  and
      Henretty, Michael  and
      Kohler, Michael  and
      Meyer, Josh  and
      Morais, Reuben  and
      Saunders, Lindsay  and
      Tyers, Francis M.  and
      Weber, Gregor",
    booktitle = "Proceedings of the 12th Conference on Language Resources and Evaluation (LREC 2020)",
    year = "2020",
    pages = "4211--4215",
}

@inproceedings{openslr65,
    title = {{Open-source Multi-speaker Speech Corpora for Building Gujarati, Kannada, Malayalam, Marathi, Tamil and Telugu Speech Synthesis Systems}},
    author = {He, Fei and Chu, Shan-Hui Cathy and Kjartansson, Oddur and Rivera, Clara and Katanova, Anna and Gutkin, Alexander and Demirsahin, Isin and Johny, Cibu and Jansche, Martin and Sarin, Supheakmungkol and Pipatsrisawat, Knot},
    booktitle = {Proceedings of The 12th Language Resources and Evaluation Conference (LREC)},
    month = may,
    year = {2020},
    address = {Marseille, France},
    publisher = {European Language Resources Association (ELRA)},
    pages = {6494--6503},
    url = {https://www.aclweb.org/anthology/2020.lrec-1.800},
    ISBN = {979-10-95546-34-4},
}

@misc{openslr127-1,
  doi = {10.48550/ARXIV.2207.13331},
  url = {https://arxiv.org/abs/2207.13331},
  author = {A, Madhavaraj and Pilar, Bharathi and G, Ramakrishnan A},
  title = {Subword Dictionary Learning and Segmentation Techniques for Automatic Speech Recognition in Tamil and Kannada},
  publisher = {arXiv},
  year = {2022},
}

@misc{openslr127-2,
  doi = {10.48550/ARXIV.2207.13333},
  url = {https://arxiv.org/abs/2207.13333},
  author = {A, Madhavaraj and Pilar, Bharathi and G, Ramakrishnan A},
  title = {Knowledge-driven Subword Grammar Modeling for Automatic Speech Recognition in Tamil and Kannada},
  publisher = {arXiv},
  year = {2022},
}

@inproceedings{wang2020multi,
    title = "Balancing Training for Multilingual Neural Machine Translation",
    author = "Wang, Xinyi  and
      Tsvetkov, Yulia  and
      Neubig, Graham",
    editor = "Jurafsky, Dan  and
      Chai, Joyce  and
      Schluter, Natalie  and
      Tetreault, Joel",
    booktitle = "Proceedings of the 58th Annual Meeting of the Association for Computational Linguistics",
    month = jul,
    year = "2020",
    address = "Online",
    publisher = "Association for Computational Linguistics",
    url = "https://aclanthology.org/2020.acl-main.754/",
    doi = "10.18653/v1/2020.acl-main.754",
    pages = "8526--8537"
}

@inproceedings{likhomanenko2021slimipl,
  title={{SlimiPL}: Language-Model-Free Data-Light Text Normalization for {ASR}},
  author={Likhomanenko, Tatiana and Xu, Qiantong and Kahn, Jacob and Synnaeve, Gabriel and Collobert, Ronan},
  booktitle={Proceedings of Interspeech},
  year={2021}
}

@INPROCEEDINGS{li2019bytes,
  author={Li, Bo and Zhang, Yu and Sainath, Tara and Wu, Yonghui and Chan, William},
  booktitle={ICASSP 2019 - 2019 IEEE International Conference on Acoustics, Speech and Signal Processing (ICASSP)}, 
  title={Bytes Are All You Need: End-to-end Multilingual Speech Recognition and Synthesis with Bytes}, 
  year={2019},
  volume={},
  number={},
  pages={5621-5625},
  doi={10.1109/ICASSP.2019.8682674}
}

@inproceedings{
    loshchilov2018decoupled,
    title={Decoupled Weight Decay Regularization},
    author={Ilya Loshchilov and Frank Hutter},
    booktitle={International Conference on Learning Representations},
    year={2019},
    url={https://openreview.net/forum?id=Bkg6RiCqY7},
}

@inproceedings{
    loshchilov2017sgdr,
    title={{SGDR}: Stochastic Gradient Descent with Warm Restarts},
    author={Ilya Loshchilov and Frank Hutter},
    booktitle={International Conference on Learning Representations},
    year={2017},
    url={https://openreview.net/forum?id=Skq89Scxx}
}

@misc{dang2026asrevalkit,
  author       = {Quy-Anh Dang},
  title        = {ASR EvalKit: A Modular Toolkit for Evaluating Automatic Speech Recognition Models},
  year         = {2026},
  url          = {https://github.com/knoveleng/asr-evalkit},
}

@inproceedings{wang2020negative,
    title = "On Negative Interference in Multilingual Models: Findings and A Meta-Learning Treatment",
    author = "Wang, Zirui  and
      Lipton, Zachary C.  and
      Tsvetkov, Yulia",
    editor = "Webber, Bonnie  and
      Cohn, Trevor  and
      He, Yulan  and
      Liu, Yang",
    booktitle = "Proceedings of the 2020 Conference on Empirical Methods in Natural Language Processing (EMNLP)",
    month = nov,
    year = "2020",
    address = "Online",
    publisher = "Association for Computational Linguistics",
    url = "https://aclanthology.org/2020.emnlp-main.359/",
    doi = "10.18653/v1/2020.emnlp-main.359",
    pages = "4438--4450"
}

@book{schiffman1999linguistic,
  title={A Reference Grammar of Spoken Tamil},
  author={Schiffman, Harold F.},
  year={1999},
  publisher={Cambridge University Press}
}

@book{deterding2007singapore,
    author = {Deterding, David},
    title = {Singapore English},
    publisher = {Edinburgh University Press},
    year = {2007},
    month = {08},
    isbn = {9780748625444},
    doi = {10.3366/edinburgh/9780748625444.001.0001},
    url = {https://doi.org/10.3366/edinburgh/9780748625444.001.0001},
}

@misc{bapna2022mslam,
      title={mSLAM: Massively multilingual joint pre-training for speech and text}, 
      author={Ankur Bapna and Colin Cherry and Yu Zhang and Ye Jia and Melvin Johnson and Yong Cheng and Simran Khanuja and Jason Riesa and Alexis Conneau},
      year={2022},
      eprint={2202.01374},
      archivePrefix={arXiv},
      primaryClass={cs.CL},
      url={https://arxiv.org/abs/2202.01374}, 
}

@misc{sitaram2019survey,
      title={A Survey of Code-switched Speech and Language Processing}, 
      author={Sunayana Sitaram and Khyathi Raghavi Chandu and Sai Krishna Rallabandi and Alan W Black},
      year={2020},
      eprint={1904.00784},
      archivePrefix={arXiv},
      primaryClass={cs.CL},
      url={https://arxiv.org/abs/1904.00784}, 
}

@inproceedings{lyu2010seame,
  title     = {{SEAME: a Mandarin-English code-switching speech corpus in south-east asia}},
  author    = {Dau-Cheng Lyu and Tien-Ping Tan and Eng Siong Chng and Haizhou Li},
  year      = {2010},
  booktitle = {{Interspeech 2010}},
  pages     = {1986--1989},
  doi       = {10.21437/Interspeech.2010-563},
  issn      = {2958-1796},
}

@inproceedings{winata2020meta,
    title = "Meta-Transfer Learning for Code-Switched Speech Recognition",
    author = "Winata, Genta Indra  and
      Cahyawijaya, Samuel  and
      Lin, Zhaojiang  and
      Liu, Zihan  and
      Xu, Peng  and
      Fung, Pascale",
    editor = "Jurafsky, Dan  and
      Chai, Joyce  and
      Schluter, Natalie  and
      Tetreault, Joel",
    booktitle = "Proceedings of the 58th Annual Meeting of the Association for Computational Linguistics",
    month = jul,
    year = "2020",
    address = "Online",
    publisher = "Association for Computational Linguistics",
    url = "https://aclanthology.org/2020.acl-main.348/",
    doi = "10.18653/v1/2020.acl-main.348",
    pages = "3770--3776"
}

@inproceedings{vaswani2017attention,
 author = {Vaswani, Ashish and Shazeer, Noam and Parmar, Niki and Uszkoreit, Jakob and Jones, Llion and Gomez, Aidan N and Kaiser, \L ukasz and Polosukhin, Illia},
 booktitle = {Advances in Neural Information Processing Systems},
 editor = {I. Guyon and U. Von Luxburg and S. Bengio and H. Wallach and R. Fergus and S. Vishwanathan and R. Garnett},
 pages = {},
 publisher = {Curran Associates, Inc.},
 title = {Attention is All you Need},
 url = {https://proceedings.neurips.cc/paper_files/paper/2017/file/3f5ee243547dee91fbd053c1c4a845aa-Paper.pdf},
 volume = {30},
 year = {2017}
}

@inproceedings{gulati2020conformer,
  title     = {{Conformer: Convolution-augmented Transformer for Speech Recognition}},
  author    = {Anmol Gulati and James Qin and Chung-Cheng Chiu and Niki Parmar and Yu Zhang and Jiahui Yu and Wei Han and Shibo Wang and Zhengdong Zhang and Yonghui Wu and Ruoming Pang},
  year      = {2020},
  booktitle = {{Interspeech 2020}},
  pages     = {5036--5040},
  doi       = {10.21437/Interspeech.2020-3015},
  issn      = {2958-1796},
}

@inproceedings{nguyen2023seallms,
    title = "{S}ea{LLM}s - Large Language Models for {S}outheast {A}sia",
    author = "Nguyen, Xuan-Phi  and
      Zhang, Wenxuan  and
      Li, Xin  and
      Aljunied, Mahani  and
      Hu, Zhiqiang  and
      Shen, Chenhui  and
      Chia, Yew Ken  and
      Li, Xingxuan  and
      Wang, Jianyu  and
      Tan, Qingyu  and
      Cheng, Liying  and
      Chen, Guanzheng  and
      Deng, Yue  and
      Yang, Sen  and
      Liu, Chaoqun  and
      Zhang, Hang  and
      Bing, Lidong",
    editor = "Cao, Yixin  and
      Feng, Yang  and
      Xiong, Deyi",
    booktitle = "Proceedings of the 62nd Annual Meeting of the Association for Computational Linguistics (Volume 3: System Demonstrations)",
    month = aug,
    year = "2024",
    address = "Bangkok, Thailand",
    publisher = "Association for Computational Linguistics",
    url = "https://aclanthology.org/2024.acl-demos.28/",
    doi = "10.18653/v1/2024.acl-demos.28",
    pages = "294--304"
}
\bibliographystyle{conference}


\appendix
\section{Dataset Details}
\label{app:datasets}
 
The following datasets were used in this work. All datasets are publicly available and used in accordance with their respective licences.

\newcolumntype{L}[1]{>{\raggedright\arraybackslash}p{#1}}
\begin{table}[ht]
\centering
\caption{Overview of datasets used in this work.}
\label{tab:dataset_overview}
\renewcommand{\arraystretch}{1.3}
\begin{tabular}{L{1.8cm} L{2.4cm} L{1.6cm} L{6.0cm} L{2.6cm}}
\toprule
\textbf{Language} & \textbf{Dataset} & \textbf{Source} & \textbf{Description} & \textbf{License} \\
\midrule
 
\multirow{2}{=}{\textbf{English}}
  & LibriSpeech \citep{panayotov2015librispeech}
  & OpenSLR
  & English read speech derived from LibriVox public-domain audiobooks.
  & CC BY 4.0 \\[4pt]
 
  & NSC$^{\dagger}$ (National Speech Corpus)
  & IMDA
  & Singapore English speech corpus covering multiple speaking styles (read, conversational, scripted).
  & Singapore Open Data Licence \\
 
\midrule
 
\multirow{4}{=}{\textbf{Mandarin}}
  & AISHELL-1 \citep{bu2017aishell}
  & OpenSLR
  & Mandarin read speech recorded by 400 speakers; ${\sim}$178 hours.
  & Apache 2.0 \\[4pt]
 
  & AISHELL-3 \citep{shi2020aishell3}
  & OpenSLR
  & Multi-speaker Mandarin TTS corpus (${\sim}$85 hours, 218 speakers) used here as ASR training data.
  & Apache 2.0 \\[4pt]
 
  & FLEURS \citep{conneau2022fleurs}
  & Google
  & Few-shot Learning Evaluation of Universal Representations of Speech; 102-language parallel benchmark.
  & CC BY 4.0 \\[4pt]
 
  & Common Voice 23
  & Mozilla
  & Multilingual crowdsourced speech; Mandarin subset from Mozilla Common Voice release 23.
  & CC0 (Public Domain) \\
 
\midrule
 
\multirow{4}{=}{\textbf{Tamil}}
  & FLEURS \citep{conneau2022fleurs}
  & Google
  & Tamil split of the FLEURS multilingual benchmark dataset.
  & CC BY 4.0 \\[4pt]
 
  & SLR127 \citep{openslr127-1, openslr127-2}
  & OpenSLR
  & IISc-MILE Tamil ASR corpus; ${\sim}$150 hours of read speech from 531 speakers.
  & CC BY 2.0 \\[4pt]
 
  & SLR65 \citep{openslr65}
  & OpenSLR
  & Crowdsourced high-quality multi-speaker Tamil speech recordings.
  & CC BY-SA 4.0 \\[4pt]
 
  & Common Voice 23
  & Mozilla
  & Tamil subset from Mozilla Common Voice release 23.
  & CC0 (Public Domain) \\
 
\midrule
 
\multirow{2}{=}{\textbf{Malay}}
  & Mesolitica\footnote{\url{https://github.com/malaysia-ai/malaysian-dataset/tree/master/text-to-speech/emilia}}
  & Mesolitica
  & Open-source Malay speech corpus for ASR research.
  & CC BY 4.0 \\[4pt]
 
  & FLEURS \citep{conneau2022fleurs}
  & Google
  & Malay split of the FLEURS multilingual benchmark dataset.
  & CC BY 4.0 \\
 
\bottomrule
\end{tabular}
\vspace{0.6em}
 
\footnotesize{$^{\dagger}$NSC is a large-scale corpus; only 100{,}000 samples (Part 1) were drawn for training in this work.}
\end{table}

\end{document}